# Leveraging Textual-Cues for Enhancing Multimodal Sentiment Analysis by Object Recognition

Sumana Biswas[1] and Karen Young[1] Josephine Griffith[1]

School of Computer Science, University of Galway, Ireland,
s.biswas2@universityofgalway.ie

**Abstract.** Multimodal sentiment analysis, which includes both image and text data, presents several challenges due to the dissimilarities in the modalities of text and image, the ambiguity of sentiment, and the complexities of contextual meaning. In this work, we experiment with finding the sentiments of image and text data, individually and in combination, on two datasets. Part of the approach introduces the novel 'Textual-Cues for Enhancing Multimodal Sentiment Analysis' (TEMSA) based on object recognition methods to address the difficulties in multimodal sentiment analysis. Specifically, we extract the names of all objects detected in an image and combine them with associated text; we call this combination of text and image data TEMS. Our results demonstrate that only TEMS improves the results when considering all the object names for the overall sentiment of multimodal data compared to individual analysis. This research contributes to advancing multimodal sentiment analysis and offers insights into the efficacy of TEMSA in combining image and text data for multimodal sentiment analysis.

## 1 Introduction

Researchers' interest in Multimodal Sentiment Analysis (MSA) has grown due to the expansion of user-generated multimodal data on social networks. MSA has become increasingly appealing and is frequently used in a wide range of fields [1], including product marketing, monitoring public opinion, and recommendations [2, 3].

Nowadays, in order to get users' attention, people often use short text and an image. Processing and analysing sentiment from single-modal data brings both opportunities and challenges. Each element alone may not fully convey its thoughts; for example, the first image and accompanying text, as shown in Table 1, are insufficient to infer sentiment on their own. On the other hand, with the development of deep learning algorithms, utilising deep convolutional neural networks encourages the joint execution of sentiment analysis tasks, known as multimodal sentiment analysis, which also introduces challenges and opportunities to the researcher. In addition, the inherent nature of these data formats is dissimilar; while text contains explicit semantic information, images primarily rely on pixel-based visual information. Moreover, the ambiguity of sentiments conveyed in both modalities can lead to differing interpretations, which further complicates the process of inferring sentiment jointly. Understanding the contextual relationship between text and the visual elements in an image necessitates the use of a technique for capturing how they work together to infer the overall or joint sentiment of multimodal data.

Most existing multimodal sentiment analysis approaches either analyse each data modality separately or concatenate the features extracted from different modalities at a coarse level. For example, Felicitte et al. (2019) [4], Zhang et al. (2019) [5], and Zhang et al. (2021) [6] analysed both visual and textual data of social media using Deep Convolution Neural Networks (DCNNs) separately as well as jointly. The authors in [4] used the visual features of the entire image, and Zhang et al. (2019) [5], and Zhanget al. (2021) [6] used the visual information of some regions of the image. Specifically, a visual image typically contains regions or objects that evoke sentiments [7]. The identified objects' names in text format can provide a textual description of the elements of the image [8–10]. In Table 1, very limited text was given; when the names of detected objects in the image are combined with the associated text with the image, they provide more information by representing both the image and the text together. For example, this list (names of detected objects) is-['tree', 'water', 'roof', 'plant', 'sky', 'sky', 'wall', 'building', 'house'] when combined with the short text-'the bucket list bora bora', gave more information together for inferring sentiment, as shown in Table 1.

In the above example, not only a few but all the object names give valuable information regarding the image, which can help infer sentiments using deep learning models. When combined with the associated caption text or superimposed text of the image data, these additional details about the image could be valuable, which was overlooked in previous work. Thus, derived textual content from the image and associated textual content may be useful when combined and used with a deep learning model to bridge the gap between these dissimilar modalities, providing a more integrated and effective solution than traditional methods that use image and text separately.

Motivated by these studies, this paper introduces the novel 'Textual-Cues for Enhancing Multimodal Sentiment Analysis' (TEMSA) based on object recognition methods. Specifically, we extract object names from the images according to [17–19]. The extracted object names are considered visual cues of the image according to [21]. At first, we identified the maximum number of objects and their names that could be recognised in the images. We then concatenated these object names with the text data, which was also available and associated with each image. We called this TEMS (Textual-Cues for Enhancing Multimodal Sentiment) as a key element of the proposed TEMSA approach, where one part of the text was purely the caption or superimposed text data of the image, and the other part came from the names of the visual objects extracted from the image.

We consider two hypotheses. First, TEMSA, an approach that combines textual information (names of detected objects) derived from images and the text data (caption or superimposed text) associated with that image, will significantly improve the performance of multimodal sentiment analysis in comparison to traditional unimodal methods that analyse image and text data separately. Second, the performance for TEMSA are improved when all of the object names are included, compared to only a single object name. We conducted four experiments: the first two experiments provide the results for comparing image-based data to text-based data individually; the third experiment used TEMS for multimodal sentiment analysis. We established the first hypothesis by conducting these first three experiments and comparing their results. For the fourth experiment, we used TEMS for multimodal sentiment analysis for a subset of each dataset, with only one object extracted from images; the results of the third and fourth experiments establish the second hypothesis. TEMSA enhances the overall inference of sentiment jointly when all detected object names are considered and compared to the results of image and text data separately on two datasets explained in Section 3.1.

**Table 1.** A few examples of multimodal data including images, caption text, or superimposed text with images and their ground truth.

| Image | 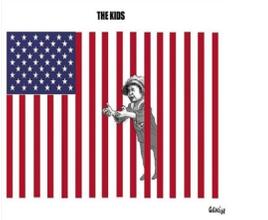 | 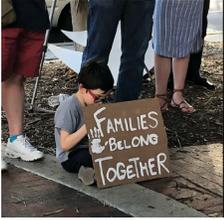 | 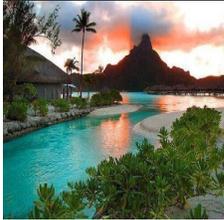 |
|---|---|---|---|
| Text | the kid genin | Families belong together | the bucket list bora bora |
| Objects Names | [man, hat, head, flag, woman, arm, person] | [person, person, hair, head, chair] | [tree, water, roof, plant, sky, sky, wall, building, house] |
| Sentiments | Negative | Neutral | Positive |

The remaining sections of the paper are as follows: Section 2 outlines previous work on sentiment analysis of multimodal data; Section 3 describes the proposed methodology that is used for the sentiment classification task; and Section 4 describes the experimental results and discussion. Section 5 concludes the paper with recommendations for future research.

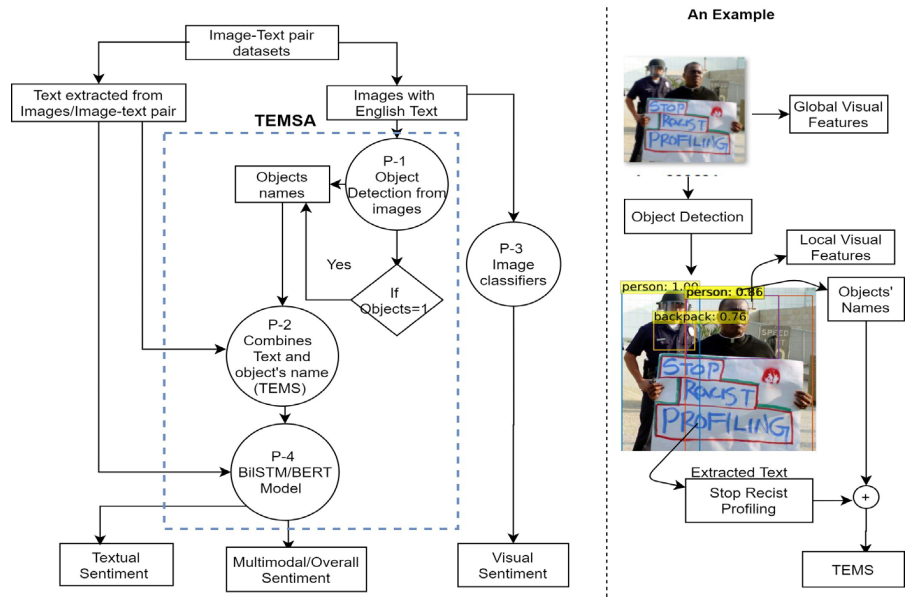

**Fig. 1.** Structure of Overall Methodology.

## 2   Previous Work

MSA has been studied for many years, and there are many effective methods that have been proposed. In this section, we review a few works from three aspects: the visual features of the entire image, which are basically abstract visual information based on whole image content, such as edges, textures, patterns, and colours; the visual features derived from the detected object in an image, which gave abstraction and visual information from the detected objects; and finally, combined methods of visual features and textual features, when any textual information is associated with the image.

The visual feature-based methods extract all the features of the entire image and construct a model to obtain visual information for visual sentiment classification [9]. Felicetti et al. (2019) [4] used Deep Convolution Neural Networks ( DCNNs) like VGG-16 net [11], AlexNet [12], CaffeNet [13], GoogLeNet [14], and ResNet [16] to extract the unimodal visual features from the entire image. Then, the DCNNs were fine-tuned to achieve the unimodal image-based sentiment analysis task. Ortis et al. (2021) [21] also employed GooglNet to extract the visual features from the whole image, which combined with the scene features of that image to infer image sentiment. Yang et al. (2018) [7] proposed a deep framework to automatically find image sentiment where they also used the visual features from the whole image.

Instead of considering all the visual information of the entire image, some researchers argue that the sentiment of an image sometimes depends on certain regions of that image [7], and the visual information of these regions can be used to better infer and predict the sentiment of the image as a whole [22]. To identify the emotions present in local regions of images, Rao et al. (2019) [23] built a region-based multi-level CNN model. They worked with 8 emotional categories and converted all emotions into 2 emotional classes (positive emotions: Amusement, Awe, Contentment, and Excitement and negative emotions: Anger, Disgust, Fear, and Sadness). Zhang et al. (2022) [20] also proposed a multilevel sentiment region correlation analysis (MSRCA) model to predict the sentiment considering the most potentially affected region in an image by emotions from multiple perspectives. They identified the filtered, relevant regional information that is potentially related to the sentiment at different levels in the original sentiment image. Yang et al. (2018) [7] proposed a deep framework to automatically find sentiment information in Affected Regions(AR) of images, which was basically object-based information. Ortis et al. (2021) [21] used textual information from the image, which provided different regional (object-based) textual descriptions. All of the literature included region-based (object-based) information for image sentiment analysis.

In contrast, Tong et al. (2022) [24] investigated the alignment between image regions and text words and proposed a cross-modal interaction between both visual and textual context information, which fails to include the remaining objects' information inferring overall sentiments. Specifically, they investigated the alignment between text word and image region, and all the detected object names may not have been present in the caption text during the alignment procedures. Moreover, sometimes no detected object name may be presented in the caption text, as we can see in the example shown in Table 1. This is why they considered only a maximum of two image regions for accounting for visual information, which reduced the sample size of the datasets. Thus, the images that have no effective region that evokes sentiment words in text or more than two effective region datasets were lost.

To avoid reducing the sample size of the dataset, Xiaojun et al. (2022) [25] and Yang et al. (2018) [7] used the visual information of one large object and the global visual information of the entire image. But they ignored the smaller objects in the image as well as the names of the most important objects. Yu et al. (2022) [26] developed three models: the BERT Text Feature Extraction model, a ResNet Image Feature Extraction model, and concatenated image and text features at the coarse level, followed by a Multimodal Feature Fusion Extraction model for multimodal sentiment analysis. The Multimodal Feature Fusion Extraction model [27] also concatenated text and image features and obtained a final multimodal prediction under the guidance of the attention mechanism. Felicetti et al. (2019) [4] used a mid-level fusion of text and image features for multimodal sentiment analysis. However, they did not incorporate the derieved object-level textual information of the images regarding sentiment.

All of the above researchers used the visual information of the whole image, part of the image, and the caption text, with each element separately and jointly, but they did not combine the most identifiable objects' names and the associated caption text for multimodal sentiment analysis.

## 3    Methodology

The proposed methodology, called TEMSA, plays a crucial role in the overall architecture as shown in Fig. 3. We utilised the two techniques of object identification depicted in Fig. 3(P1). Firstly, we used DETR (End-to-End Object Detection with Transformers), which was optimised for RCNN baseline on the challenging Coconames datasets, developed by Carion et al. (2020) [17] to identify the maximum number of objects for a given input image. Secondly, we used Faster R-CNN from [28], pre-trained on the Visual Genome dataset from [18] using ResNet101 according to [16] as a backbone. When the top 200 categories of object names from the Visual Genome dataset are considered, only 91 categories of object names from the Coconames dataset are taken into account [18, 29–31]. However, we combined all the objects' names extracted from the two object detection algorithms, and then we concatenated these objects' names with the available text data associated with an image, as shown in Fig. 3, P2. We refer to this as the TEMS, a key component of TEMSA, of an image, where one part of the text is made up purely of the text data associated with the image, and the other portion is derived from the names of the image's visual objects.

We conducted four sentiment analysis experiments. In the first experiment, we used the entire image data—using only the visual data—to infer unimodal visual-based sentiment using DCNNs [4] models, and Vision Transformer (ViT) [32]. The second experiment used the BiLSTM [33] and BERT [35] models for unimodal text-based sentiment analysis using text data. The TEMS was used in the BiLSTM and BERT models to obtain the overall sentiment in the third experiment. In the fourth experiment, we used the TEMS for a subset of the whole dataset that had only one object detected by the DETR method [17]. The aim of this experiment was to understand the significance of using TEMS with one object to determine overall sentiments.

**Table 2.** Summary statistics of two datasets

| Datasets | SIMPSoN | | | | MVSA-Single | | | |
|---|---|---|---|---|---|---|---|---|
| Sentiment labels | Pos | Neg | Neut | Total | Pos | Neg | Neut | Total |
| Image_labels | 1773 | 2112 | 5362 | 9247 | 2628 | 1185 | 888 | 4071 |
| Text_labels | 1008 | 745 | 1077 | 2830 | 1836 | 1687 | 1178 | 4071 |
| Joint_labels | 1016 | 798 | 1016 | 2830 | 1371 | 704 | 411 | 2486 |

### 3.1 Datasets

We evaluated the experiments on two datasets collected from two renowned social media platforms: Instagram and Twitter.

**SIMPSoN Dataset** The first dataset used is the SIMPSoN dataset, which was created by Felicetti et al. [4] to analyse the image, text, and overall sentiment of social media images using Deep Convolution Neural Networks (DCNNs). In the dataset, daily news-related images were collected from Instagram, and the sentiment of 9247 images was manually labelled. Every image was labelled for sentiment regarding only its visual (Image_labels), only its textual(Text_labels), and then for both the image and text (Joint_labels). One of three sentiments: "positive", "negative", and "neutral" were assigned in each of the cases. Images displaying solidarity, friendliness, and, in general, all positive facts were assigned a "positive" sentiment; images depicting violence, racism, and overly vulgar comments were assigned a "negative" sentiment; otherwise, the sentiment assigned was "neutral". Since some of the images have no text and some of them have non-English text, we considered a subset of the dataset containing the images that had superimposed text information for text and joint sentiment analysis. After removing the images with non-English text, we obtained 2830 examples.

**MVSA Single** The second dataset used is MVSA Single, where originally 5,129 image-text pairs were collected from Twitter. Each pair had a sentiment -— positive, neutral, or negative —- for the text (Text labels) and for the image (Image_labels); the annotation method was described in [36]. After pre-processing, 4071 images were obtained. Since this dataset had no overall sentiment labels for the joint text and image and our objective is to find similar sentiments for both modalities, we created a new Joint_label for the overall sentiment of text and image, departing from the previous approaches in [37, 24]. First, we removed images where the image labels were positive while the text labels were negative, and vice versa. For any remaining images, the overall sentiment was taken as the same sentiment label given to both the image and text. After this, we had 2486 images with the new Joint_labels for the MVSA-Single dataset. Table 2 contains the statistics of the datasets.

**Table 3.** The percentage of images for both datasets that have different numbers of identified objects.

| Number of Identified Objects | 0 | 1 | 2 | 3 | 4 | 5 | 6 | 7 |
|---|---|---|---|---|---|---|---|---|
| SIMPSoN | 36% | 28% | 18% | 9% | 6% | 4% | 1% | 0.03% |
| MVSA-Single | 15% | 22% | 16% | 10% | 0.07% | 0.04% | 0.03% | 0.02% |

### 3.2   Analysis of Recognised Objects

Table 3 provides a partial view of the percentage of images that have a certain number of objects identified, i.e., some (36% in the SIMPSoN dataset) had no identifiable objects, and some of the images had 45 objects identified. The issue of different numbers of objects per image is a very important and challenging one in incorporating visual and textual information jointly in the deep learning model. To handle this challenge, we analysed the entire dataset with respect to most of the objects detected in each image. In contrast, we conducted one experiment with a subset of the dataset images with only one object. This experimental subset consists of 28% of the SIMPSoN dataset and 22% of the MVSA Single dataset, indicating that it contains the greatest amount of data involving a single object, as shown in Table 2. The purpose of this experiment is to determine the effect of considering only single-object images to determine overall sentiments.

### 3.3   Pre-processing Text Data and Extracting TEMS

For the text data, we removed all non-alphanumeric symbols, emails, and tags, and changed the text to lowercase. Then, the text data was tokenised and transformed into vector sequences. We used a pre-trained BERT-Base [35] model to embed each word into a 768-dimensional embedding vector and a 300-dimensional embedding vector for GloVe [39]. In the same way, the TEMS data was also cleaned and tokenised.

### 3.4   Experimental Parameter Settings and Model Structure

The datasets are divided into 80% training and 20% testing according to the 8:2 ratio, according to [4]. The model's structure and its hyperparameter settings are described below.

**Models for Image Data** For the first experiment, we used the image data. Each image was resized to 224*224 pixels. Each image was randomly flipped left-to-right, top-to-bottom, and rotated (90/180/270) to improve the training datasets to obtain visual (image-based) sentiment. We used three baseline DCNNs models: VGG16, VGG19, and RESNET50 which were pre-trained on the ImageNet2K [38] dataset (see "P3" in Fig. 3). Without training this model, we cut the final classification layer and added one dense layer with 1024 hidden units with a Relu activation function, and one dropout layer followed by a Softmax activation function to obtain the final classification of visual sentiments. We used the activation, a Gelu, for ViT, and one dropout layer followed by a Softmax activation function to obtain the final classification of visual sentiments. All learning rates were 8e-4 for all models, and we used the image_labels, as summarised in Table 2, to train and test the images for the two datasets.

**Models for Text and TEMS** We used BILSTM and BERT models (see "P4" in Fig. 3) for unimodal text-based sentiment classification using purely text data and TEMS for overall (multimodal) sentiment classification.

**BiLSTM Model** Bidirectional Long Short-Term Memory (BiLSTM) is a Recurrent Neural Network (RNN) with three gates: input gate, output gate, and forgot gate [33]. We have used the BiLSTM layer with 32 hidden units followed by one dropout layer and a Softmax activation function with three classes of probability. One hot encoding technique was used to deal with the categorical data, which creates a binary column for each category (and thus we also use the Softmax function for the final sentiment classification). BiLSTM model, considering each of the text-initialised words vectors T = ($w_1$, $w_2$, $w_3$, ..., $w_m$) in one T (here only text data or TEMS) as the input of the BiLSTM networks, the output of the BiLSTM hidden layer contain every context's forward output and backward output ($\overleftarrow{c}_i$), concatenating forward ($\overrightarrow{h}_i$) and backward output ($\overleftarrow{h}_i$) the context is expressed as follows, which are summarised from [34, 33]:

$$\overleftarrow{c}_i = \left[\overleftarrow{h}_i; \overrightarrow{h}_i\right]^L, i = 1, \ldots,$$

$$\overleftarrow{h}_i = \text{LSTM}_l\left(\overleftarrow{h}_{i-1}, \overleftarrow{w}_i\right),$$

$$\overrightarrow{h}_i = \text{LSTM}_r\left(\overrightarrow{h}_{i+1}, \overrightarrow{w}_i\right).$$

Here, $LSTM_l$ and $LSTM_r$ are the LSTM of the left and right direction, respectively.

**BERT Model** Bidirectional Encoder Representations from Transformers (BERT) is a transformer-based machine learning technique. It was developed and released in 2018 by Google's Jacob Devlin and his coworkers [35] for natural language processing (NLP) pre-training. The models use pre-trained unlabeled data corpus of 800 million words and the English Wikipedia corpus of 2.5 billion words, derived from the BooksCorpus [40]. There are primarily three layers. The BERT Input representation layer consists of a sequence of tokens, including tokens for sentence separation and classification. An embedding layer transforms these tokens into vectors for embedding. Transformer Encoder Layers are comprised of multiple transformer encoder layers; we considered 12 of them, each containing a set of self-attention and feed-forward neural network sub-layers [35]. Lastly, self-attention mechanisms enable BERT to simultaneously assess the relationships between all words in the input sequence. Thus, the detected objects' names make a relation with the caption or superimposed text. It computes attention weights for each word in the sequence based on the proximity of other words to the sentiment-related (in our case) target word.

We added one Dense layer with 1024 hidden units and The Relu activation function followed by one dropout layer and a Softmax activation function for the final sentiment classification. The mathematical representations are given as follows for the self-attention mechanism, multiheaded-self attention mechanism, and feed-forward network mechanism for the final output respectively, for a given word, denoted as "i", in the input sequence of the whole text, "T"; The derivations and equations below are summarised from [35, 34].

$$\text{Attention}(Q_i, K_j, V_j) = \sum_{j=1}^{T} \frac{\exp(Q_i \cdot K_j)}{\exp(Q_i \cdot K_j)} \cdot V_j$$

Where $Q_i$ is the query vector for the word "i",
$K_j$ and $V_j$ are the key and value vectors for the word "j",
T is the total number of words in the input sequence.

$$\text{Output}_{\text{self-attention}} = \text{LayerNorm}(\text{Input} + \text{MultiHeadAttention}(\text{Input}))$$

$$\text{Output}_{\text{FFN}} = \text{LayerNorm}(\text{Output}_{\text{self-attention}} + \text{FeedForward}(\text{Output}_{\text{self-attention}}))$$

For experiment 2, we used the text_labels, as summarised in Table 2, to train and test the pure text data for the two datasets. The number of words in each phrase varied in the SIMPSoN dataset; this is why we took a maximum of 55 words, truncating long phrases and padding the short phrases with Zero values according to [4]. The maximum number of words was 21 for the MVSA-Single dataset, where Zeros were padded for shorter phrases [4].

For experiment 3, we used the joint_labels (image and text), as shown in Table 2, to train and test with TEMS for the two datasets. The word length of the TEMS was increased after combining objects' names with text data; we set the maximum number of words to 75 for the SIMPSoN and 41 for the MVSA Single, allowing for a maximum of 20 object names according to [43]. Zeros were padded for less than 20 object names.

The proposed models were optimised using the Adam optimiser. We set the batch size for each dataset to 32, and we run 10 epochs for all the experiments according to [6]. The learning rates were 1e-02 and 6e-06 for the BiLSTM and BERT models, respectively. In the training process, to avoid overfitting, we used a dropout rate of 0.1 for the two models.

**Table 4.** Experiment 1: Results of unimodal image-based sentiment classification.

| Model | SIMPSoN(%) | | | | MVSA-Single(%) | | | |
|---|---|---|---|---|---|---|---|---|
| | Acc | Pre | F1 | Rec | Acc | Pre | F1 | Rec |
| Vgg16 [4](BL) | 74 | 72 | 73 | 74 | 54 | 46 | 46 | 45 |
| RsNet50 [4](BL) | 73 | 69 | 70 | 71 | 59 | 56 | 54 | 55 |
| Vgg19 | 73 | 71 | 73 | 73 | 49 | 45 | 46 | 45 |
| ViT | 58 | 49 | 50 | 47 | 47 | 34 | 38 | 27 |

## 4  Experimental Results and Discussion

In this section, we present the results of the four experiments. The results are reported using accuracy, precision, recall, and F1 scores. We compare the results of the baseline (BL) methods for image-based data [4] of the first experiment, and text-based data [41, 42] of the second experiment. The best results obtained with TEMS are highlighted in bold in the third experiment. We compare the results of our proposed approach for multimodal sentiment analysis to the results obtained for image-based (visual), and text-based (textual) sentiment analysis.

**Table 5.** Experiment 2: Results of unimodal text-based sentiment classification.

| Model | SIMPSoN(%) | | | | MVSA-Single(%) | | | |
|---|---|---|---|---|---|---|---|---|
|  | Acc | Pre | F1 | Rec | Acc | Pre | F1 | Rec |
| Zhang [41] (BL) | 57 | 54 | 52 | 53 | 58 | 57 | 58 | 55 |
| Kim [42] (BL) | 68 | 67 | 63 | 63 | 63 | 64 | 64 | 62 |
| BiLSTM | 65 | 64 | 51 | 56 | 64 | 69 | 65 | 62 |
| BERT | 69 | 66 | 65 | 65 | 70 | 69 | 69 | 69 |

**Table 6.** Experiment 3: Results of TEMS for overall (multimodal) sentiment classification.

| Model | SIMPSoN(%) | | | | MVSA-Single(%) | | | |
|---|---|---|---|---|---|---|---|---|
|  | Acc | Pre | F1 | Rec | Acc | Pre | F1 | Rec |
| BiLSTM | 67 | 61 | 57 | 59 | 66 | 70 | 66 | 64 |
| BERT | **79** | **74** | **75** | **74** | **84** | **78** | **78** | **79** |

We can see Vgg16 obtained the highest result in the first experiment for both datasets using the baseline image-based data (visual features from the entire image), in the results shown in Table 4 as like as [4]. The Vision Transformer (ViT) did not perform well in the sentiment analysis task. The results did not improve compared to the baseline models.

Table 5 depicts the results of the second experiment for textual sentiment classification. The performance of the BiLSTM model gave a slightly poorer result when textual features were extracted by GloVe. The pre-trained language model BERT helped to improve the performance of textual sentiment classification and the results have been improved compared to the baseline models of the previous methods used [41, 42].

The result of the proposed method of TEMS using the BERT model obtained the best result for both datasets in the third experiment (Table 6). In addition, when TEMS is used in the BERT model, the mechanism of the BERT model has been advantageous in bridging the gap between word-to-word relations and detecting target sentiment. Especially, when image and text data jointly provide overall (jointly) sentiment for multimodal data. Additionally, the results of the Wilcoxon signed-rank test [44] with alpha values set at 0.05 revealed a statistically significant difference after comparing the sentiment labels of Joint_labels and predicted labels of the third experiment using TEMS for both datasets.

The results of the fourth experiment are presented in Table 7. Here, we notice that the result of TEMS was inferior when only one object class name was included, which is the likely reason for the inferior result.

**Table 7.** Experiment 4: Results of TEMS with one object name for overall sentiment classification.

| Model | SIMPSoN(%) | | | | MVSA-Single(%) | | | |
|---|---|---|---|---|---|---|---|---|
| | Acc | Pre | F1 | Rec | Acc | Pre | F1 | Rec |
| BiLSTM | 61 | 58 | 50 | 55 | 63 | 60 | 61 | 60 |
| BERT | 62 | 62 | 59 | 60 | 69 | 66 | 63 | 64 |

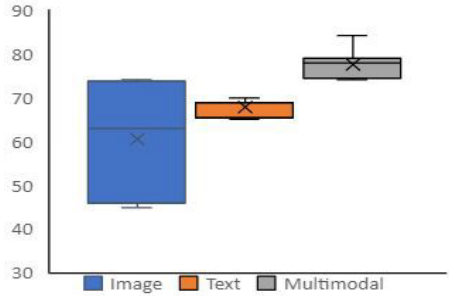

**Fig. 2.** Result Comparison of text, image and multimodal sentiment, and '×' indicates the mean performance.

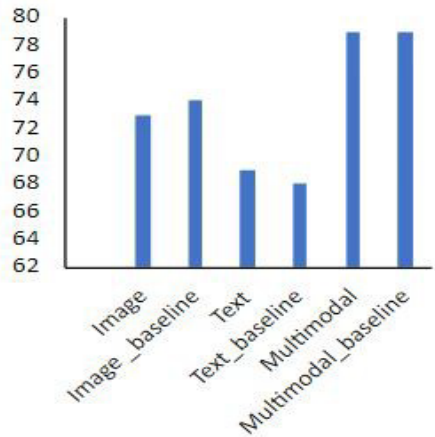

**Fig. 3.** Result comparison of text, image and multimodal sentiment with baseline for SIMPSoN datasets.

From the above results, we see that the results of TEMS in (Table 6) are superior to all the results (Table 4) which were produced using only image data, and all the results (Table 5) which were produced using only text data. We believe that the TEMS, that is, the concatenation of text and object names, helped to improve the results over using image-based and text-based data individually. Fig. 2 shows the mean comparison of image, text, and multimodal sentiment, where '×' indicates the mean performance. We also observe that the proposed TEMS approach achieved the best results for the two datasets but only when more than one object name was included. The TEMS is replete with all detected visual information in an image, including the names of all objects, and therefore includes information on the visual objects in text form, which addresses the issues of data dissimilarity in terms of data format as well as data types (all are text data). The results of TEMS show similar accuracy with [4] for the SIMP-SoN dataset as shown in Fig. 3. The results of TEMS were very good for MVSA Single datasets. This is the first time we have trained and tested the target sentiment (joint labels), departing from the previous work [37, 24], in which both the image and text had similar sentiment labels for the MVSA Single dataset using this methodology, making it unique with no previous comparisons. From the above observations, we can conclude that each experiment was important to strengthen the rationality of the proposed TEMS approach to analyse overall sentiment.

## 5   Conclusion

In conclusion, the TEMSA approach emerges as a valuable approach to achieving multimodal sentiment for image and text data, which can enhance the utility of MSA in various real-world applications. Using the robust BERT model for sentiment detection through textual cues (TEMS) from image and text data has yielded substantial benefits, as demonstrated by the outcomes of four experiments on two datasets. We obtained the best result using only the TEMS when including all object names. Thus, the authors argue that this approach is straightforward, and all experimental results demonstrate the significance and rationality of the proposed method.

We conducted all the experiments on relatively small datasets, which must be acknowledged. We plan to extend our research to include more visual information about the image, thereby facilitating multimodal sentiment analysis on large and more diverse datasets. In addition, we intend to investigate more effective techniques for extracting features from images, such as face detection, facial expression analysis, and scene comprehension; combining these features with the available text will undoubtedly enable future improvements to multimodal sentiment analysis.

16      Sumana Biswas et al.38. Ridnik, T., Ben-Baruch, E., Noy, A., Zelnik-Manor, L.: Imagenet-21k pretraining for the masses. arXiv preprint arXiv:2104.10972 (2021)
39. Pennington, J., Socher, R., Manning, C.D.: Glove: Global vectors for word representation. In: Proceedings of the 2014 conference on empirical methods in natural language processing (EMNLP). pp. 1532–1543 (2014)
40. Yukun Zhu, Ryan Kiros, Rich Zemel, Ruslan Salakhutdinov, Raquel Urtasun, Antonio Torralba, and Sanja Fidler. Aligning books and movies: Towards story-like visual explanations by watching movies and reading books. In Proceedings of the IEEE international conference on computer vision, pages 19–27 (2015)
41. Zhang, X., Zhao, J., LeCun, Y.: Character-level convolutional networks for text classification. Advances in neural information processing systems 28 (2015).
42. Kim, Y., Jernite, Y., Sontag, D., Rush, A.: Character-aware neural language models. In: Proceedings of the AAAI conference on artificial intelligence. vol. 30 (2016).
43. Biswas, S., Young, K., and Griffith, J. (2023). Automatic sentiment labelling of multi- modal data. In Cuzzocrea, A., Gusikhin, O., Hammoudi, S., and Quix, C., editors, Data Management Technologies and Applications, pages 154–175, Cham. Springer Nature Switzerland.
44. Benavoli A, Corani G, Mangili F. Should we really use posthoc tests based on meanranks? The Journal of Machine Learning Research.,17(1):152–161 (2016)